\documentclass[10pt,twocolumn,letterpaper]{article}

\usepackage{iccv}
\usepackage{times}
\usepackage{epsfig}
\usepackage{graphicx}
\usepackage{amsmath}
\usepackage{amssymb}
\usepackage{verbatim}
\usepackage{bbm}
\usepackage{subcaption}
\usepackage{booktabs}
\usepackage{lipsum}  

\usepackage[pagebackref=true,breaklinks=true,letterpaper=true,colorlinks,bookmarks=false]{hyperref}

\iccvfinalcopy 

\def\iccvPaperID{3738} 

\ificcvfinal\fi

\begin{document}

\title{Cerberus: A Multi-headed Derenderer}

\author{Boyang Deng\thanks{Work done as part of the Google AI Residency Program}, Simon Kornblith, and Geoffrey Hinton\\
Google Research, Brain Team\\
{\tt\small \{bydeng, skornblith, geoffhinton\}@google.com}
}

\maketitle

\begin{abstract}
To generalize to novel visual scenes with new viewpoints and new object poses, a visual system needs  representations of the shapes of the parts of an object that are invariant to changes in viewpoint or pose.

3D graphics representations disentangle visual factors such as viewpoints and lighting from object structure in a natural way.
It is possible to learn to invert the process that converts 3D graphics representations into 2D images, provided the 3D graphics representations are available as labels. When only the unlabeled images are available, however, learning to derender is much harder. 

We consider a simple model which is just a set of free floating parts. Each part has its own relation to the camera and its own triangular mesh which can be deformed to model the shape of the part. At test time, a neural network looks at a single image and extracts the shapes of the parts and their relations to the camera. Each part can be viewed as one head of a multi-headed derenderer. During training, the extracted parts are used as input to a differentiable 3D renderer and the reconstruction error is backpropagated to train the neural net. We make the learning task easier by encouraging the deformations of the part meshes to be invariant to changes in viewpoint and invariant to the changes in the relative positions of the parts that occur when the pose of an articulated body changes.   

Cerberus, our multi-headed derenderer, outperforms previous methods for extracting 3D parts from single images without part annotations, and it does quite well at extracting natural parts of human figures.
\end{abstract}

\section{Introduction}
\label{sec:intro}

\begin{figure}[t]
  \centering
    \begin{subfigure}[t]{.45\linewidth}
      \centering
      \includegraphics[width=0.95\linewidth]{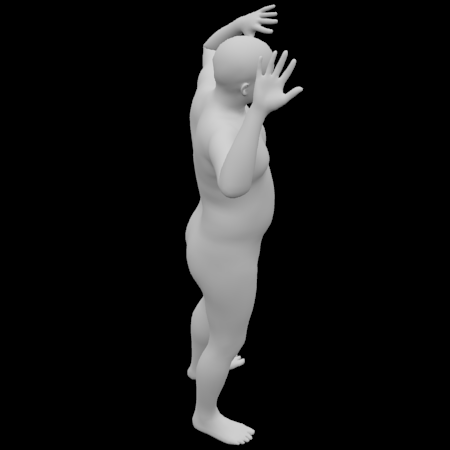}
      \caption{Input}
      \label{fig:fig1intput}
    \end{subfigure}
    \hspace{.05\linewidth}
    \begin{subfigure}[t]{.45\linewidth}
      \centering
      \includegraphics[width=0.95\linewidth]{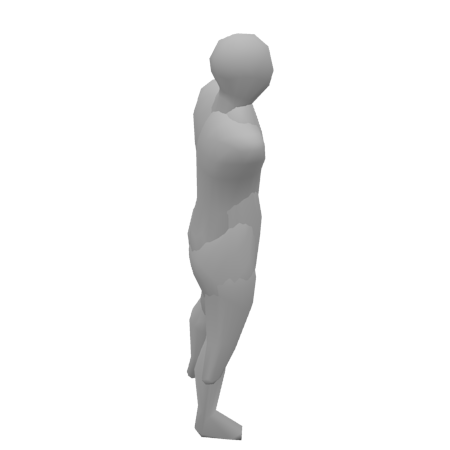}
      \caption{3D Output}
      \label{fig:fig1output}
    \end{subfigure}
    \begin{subfigure}[t]{.45\linewidth}
      \centering
      \includegraphics[width=0.95\linewidth]{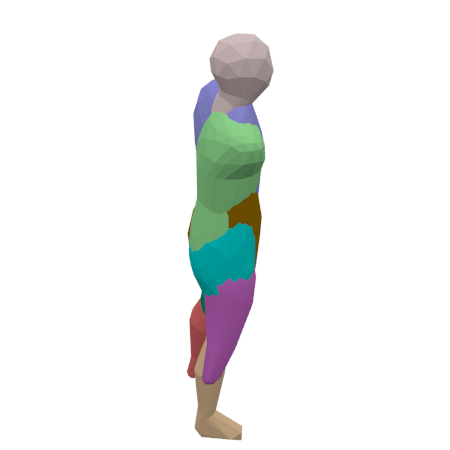}
      \caption{Parts}
      \label{fig:fig1parts}
    \end{subfigure}
    \hspace{.05\linewidth}
    \begin{subfigure}[t]{.45\linewidth}
      \centering
      \includegraphics[width=0.95\linewidth]{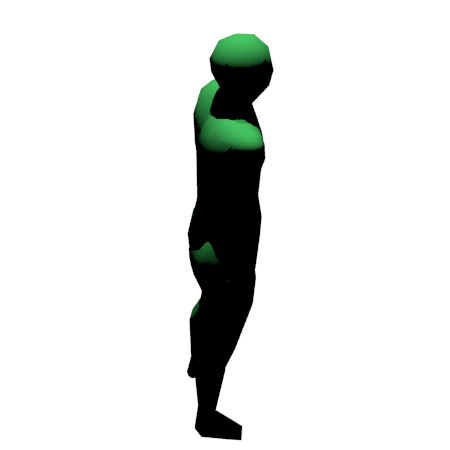}
      \caption{New Lighting}
      \label{fig:fig1newlight}
    \end{subfigure}
    \begin{subfigure}[t]{.45\linewidth}
      \centering
      \includegraphics[width=0.95\linewidth]{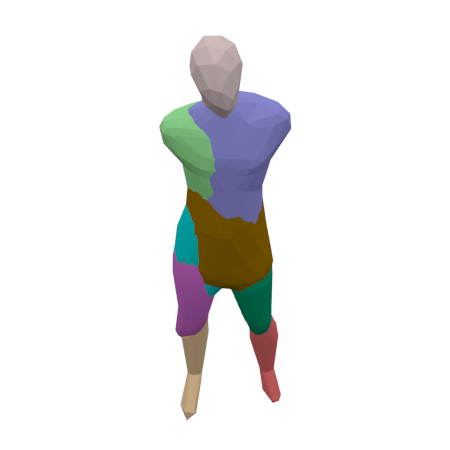}
      \caption{New Viewpoint}
      \label{fig:fig1newvp}
    \end{subfigure}
    \hspace{.05\linewidth}
    \begin{subfigure}[t]{.45\linewidth}
      \centering
      \includegraphics[width=0.95\linewidth]{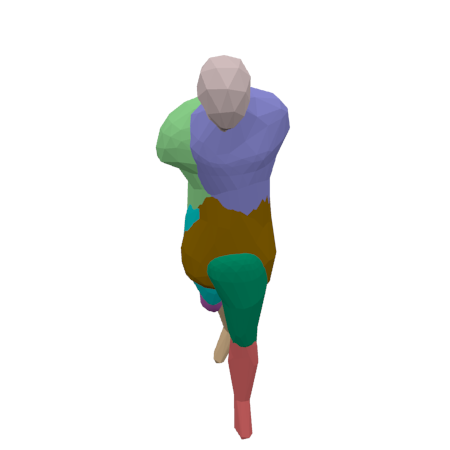}
      \caption{New Pose}
      \label{fig:fig1newpose}
    \end{subfigure}
  \caption{Given an input image (\subref{fig:fig1intput}), Cerberus can output a 3D model of the object~(\subref{fig:fig1output}). This 3D model has multiple parts (shown in different colors in~\subref{fig:fig1parts}). With this 3D model, we can render images with new lighting~(\subref{fig:fig1newlight}) or from a new viewpoint~(\subref{fig:fig1newvp}). We can also manipulate the parts and generate a new pose~(\subref{fig:fig1newpose}).}
  \label{fig:generalize}
\end{figure}

Over the years, many efforts have been made to learn visual representations by reconstructing inputs in pixel space~\cite{turk1991face, hinton2006reducing, kingma2013auto}. Empirically, learned models have smooth latent manifolds and are able to generate outputs that resemble natural images.
However, each pixel is affected by multiple factors, \eg lighting, viewpoint, surface reflectance, and surface shape. Modeling pixel responses directly is challenging when the generative model that combines these factors must be learned from data.
By contrast, in the computer graphics community, techniques that use disentangled representations are well-established. The process of generating pixels from these representations, \ie rendering, has also been exhaustively explored.
Given a representation similar to the one used in graphics, it is straightforward to generalize to new viewpoints, new poses of the objects, or new lighting conditions (as shown in Figure~\ref{fig:generalize}).
Moreover, with the advent of differentiable renderers~\cite{loper2014opendr, kato2018neural, li2018differentiable,henderson2018learning}, we can avoid the expensive acquisition of 3D labels and learn 3D graphics representations simply by reconstructing 2D images and backpropagating the reconstruction error.

In the graphics world, complex objects are modeled by dividing them into simple parts.
Decomposition into parts is especially important for applications such as gesture recognition and augmented reality which must deal with 
articulated bodies that can adopt a wide range of poses.
Although there are clear benefits to part-based models, learning natural parts without the benefit of part annotations is difficult.
Previous work has demonstrated that it is possible to learn sparse part information such as keypoints~\cite{suwajanakorn2018discovery, jakab2018unsupervised} without requiring any supervision by making use of the way images transform. Here, we seek to discover dense part information without requiring part annotations.

To this end, we present Cerberus, a neural network that extracts a part-based 3D graphics representation from a single image, along with a training strategy that avoids the need for part annotations by using natural consistencies, \ie the invariance of part shapes under changes in viewpoint or pose.\footnote{In work dealing only with rigid bodies, the word {\em pose} is often used to refer to the position and orientation of the object relative to the camera. In this work, we use pose to refer to the relative positions and orientations of the parts of an articulated body, as in human pose estimation.}
This training strategy ensures that Cerberus can learn to reconstruct geometrically correct 3D graphics models consisting of semantic parts without part supervision.
The arrangements of the 3D parts extracted by Cerberus change with pose, and we can manipulate the 3D model to form a novel pose (Figure~\ref{fig:fig1newpose}).

We examine Cerberus on two datasets of articulated bodies.
On the human dataset, which has substantial variability in pose, Cerberus not only outperforms previous work by a large margin, but also learns semantic parts such as head and legs without part annotations. These parts are consistent across poses: Cerberus produces better results than baselines even when it is restricted to applying parts extracted from an image of an individual to all other images.

In this work, we introduce the problem of unsupervised 3D perception of articulated bodies with only 2D supervision. Our key contributions are as follows:
\begin{itemize}
  \item We propose a new architecture, Cerberus, for single image 3D perception. This architecture is more suitable for modeling articulated bodies than previous architectures.
  \item We tackle the problem of learning semantic parts without part supervision by using natural but powerful consistency constraints.
  \item Our architecture, trained with the proposed constraints, outperforms baselines, even when we restrict it to extract one set of parts and apply to use the same parts for all configurations of the same subject.
\end{itemize}
\begin{figure*}[t]
  \begin{center}
    \includegraphics[width=0.95\linewidth]{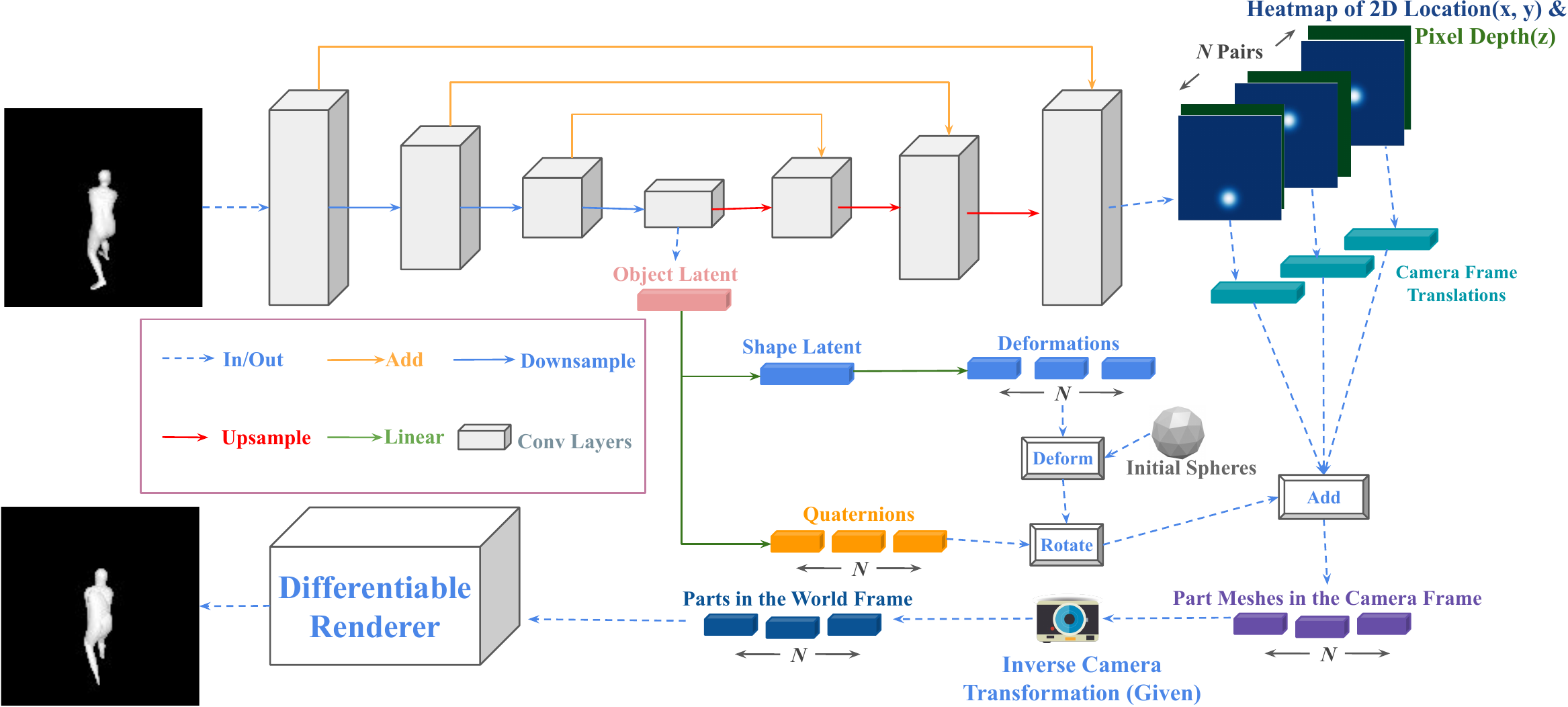}
  \end{center}
  \caption{Cerberus architecture. Here we visualize $3$ out of $N$ parts used in the pipeline. Up-sampling is performed by de-convolution. The object latent is obtained by global average pooling of the lowest-resolution feature maps. Predicted quaternions are used to construct rotation matrices.}
  \label{fig:pipeline}
\end{figure*}

\section{Related Work}
\label{sec:realted}
The idea of computer vision as inverse graphics has a long history \cite{roberts1963machine,baumgart1974geometric,grenander1978pattern,mumford1994pattern}. Recent approaches claiming to perform inverse graphics typically consist of an encoder and decoder, with a latent space that has some meaning relative to underlying generative factors. Transforming autoencoders \cite{hinton2011transforming} model images by factorizing them as a set of {\em capsules} with corresponding 3D pose vectors, such that applying a 3D rotation to the 3D pose vector produces a rotated output. 
Other work has clamped latent variables to align to generative factors \cite{kulkarni2015deep} or imposed information constraints on the latent space \cite{higgins2017beta}. We instead prespecify the form of the latent representation by using a fixed differentiable renderer as the decoder. This strategy is common in recent work that learns 3D representations \cite{loper2014opendr, kato2018neural,li2018differentiable,henderson2018learning,kanazawa2018learning}. In 2D, Tieleman~\cite{tieleman2014optimizing} also used a fixed decoder, reconstructing images based on affinely transformed learned templates.

We recover a 3D graphics representation from a single image, using only 2D supervision during training. Many previous approaches to inferring 3D representations have employed 3D supervision  \cite{choy20163d,wu2016learning,tatarchenko2017octree,fan2017point,wang2018pixel2mesh,richter2018matryoshka} or fit low-dimensional parameterized models \cite{bogo2016keep,zuffi2018lions,Genova_2018_CVPR}. Nonetheless, there is a significant body of previous work that has used only 2D supervision. Non-neural network-based approaches have reconstructed 3D models from segmentation masks by combining SfM viewpoint estimation with voxel-based visual hull \cite{vicente2014reconstructing} and deformable point cloud approaches \cite{kar2015category}. Neural network-based approaches have inferred 3D models using perspective projection \cite{yan2016perspective} or ray-tracing \cite{tulsiani2017multi,tulsiani2018multi} of volumetric representations, differentiable point clouds \cite{insafutdinov2018unsupervised}, prediction of multiple 2.5D surfaces \cite{shin2018pixels}, REINFORCE gradients through off-the-shelf renderers \cite{rezende2016unsupervised}, or fully differentiable mesh renderers \cite{kato2018neural,henderson2018learning,kanazawa2018learning}.

Part-based 3D perception and modeling has a rich past, although it has recently fallen somewhat out of favor. Early computer vision projects attempted to develop programs capable of recognizing compound objects as combinations of parts~\cite{roberts1963machine,guzman1968decomposition}. Biederman~\cite{biederman1987recognition} influentially suggested that human object perception operates by decomposing objects into 36 primitive generalized-cones (geons). More recently, van den Hengel \etal~\cite{van2015part} proposed a method to estimate the constituent parts and arrangement of Lego models based on multiple silhouettes. AIR~\cite{eslami2016attend} infers arrangements of prespecified meshes using a 3D renderer using finite-difference gradients. Other work has attempted to model 3D volumes with fixed primitives, obtaining plausible object parsings without explicit part-level supervision \cite{tulsiani2017learning,zou20173d,tian2018learning}. In contrast to these approaches, we learn rich part shapes in addition to positions, which allows us to extract the complex surface shapes of articulated bodies. 
\section{Cerberus Architecture}
\label{sec:architecture}

\subsection{3D Parameterization}
\label{sec:3dparam}

Polygonal meshes, which are widely used in computer graphics, are an effective way to define shapes. A polygonal mesh is a collection of vertices, edges and faces that together describe the surface of an object. Each vertex can also be associated with auxiliary attributes like texture and albedo.
Compared with voxel representations used by some previous works~\cite{yan2016perspective, wu2016learning}, polygonal meshes are a more compact 3D representation and are easier to render with complex shading.
In this work, we use triangular mesh (referred as "mesh" for the rest of this paper) as our 3D representation.
One challenge of using meshes for learning 3D shapes is predicting the correct connectivities between vertices. Without constraints, neural networks are prone to erroneous connectivity predictions.
To overcome this issue, we construct meshes by deforming a spherical mesh \cite{kato2018neural}.
The edges, faces, and initial positions of all the vertices of this sphere are predefined. To model the shape of a part, the neural network predicts only the displacement of each vertex.

Since Cerberus models articulated objects, we seek to develop a representation that can be manipulated to allow the modeled object to take on different poses.
Using a one-piece mesh for an articulated object makes this challenging because of the difficulty of finding vertex-wise correspondence between poses. In contrast, a part-based model can easily be made to pose in various ways.

Here we use an independent mesh for each part. We parameterize a part's local pose by its rotation and translation relative to the camera. The neural network predicts parameters of these transformations based on the pose of the object in the input image.
After applying transformations to each part and putting them together in the same 3D space, we obtain a render-ready 3D model of the whole object in a specific pose.

\subsection{3D Reconstruction Pipeline}
\label{sec:3dpipeline}

Our pipeline is illustrated in Figure~\ref{fig:pipeline}.
Given an input image, our pipeline outputs the 3D parameters defined in Section~\ref{sec:3dparam} for all the parts. The number of parts, $N$, is a predefined hyper-parameter. As shown in Figure~\ref{fig:pipeline}, we use a base network similar to the hourglass block~\cite{newell2016stacked} to extract deformation, rotation, and translation parameters from a single image. We describe the process to get each of these parameters below.

\paragraph{Deformation:}We predict vertex deformations of all the parts simultaneously. The input image goes through a down-sampling neural network to produce lower-resolution feature maps. We use global average pooling to transform these feature maps into a feature vector, referred to as the \textit{object latent} in Figure~\ref{fig:pipeline}. We linearly transform this object latent to get a shape latent, and then again linearly transform this shape latent to yield the vertex deformations. The shape latent is used to disentangle deformation (shape) and rotation (pose) as well as to implement pose consistency (described in Section~\ref{sec:posecons}).

\paragraph{Rotation:}We use quaternions for rotations. A linear transformation is performed on the object latent to produce quaternions. We construct rotation matrices based on quaternions and multiply them with the corresponding deformed parts.

\paragraph{Translation:}Instead of predicting translation parameters directly, we retrieve 3D translations from 2D coordinates and depth using an approach similar to KeypointNet~\cite{suwajanakorn2018discovery}. After the down-sampling network, we use an up-sampling network with skip connections to enlarge feature maps to the same resolution as the input.
For each part, we linearly transform the feature maps and apply a spatial softmax, yielding a ``probability map" $\{p_{x,y}^k\}$. 
We compute the 2D coordinates for the part by taking the expectation over this map. We also calculate a depth map $\{d_{x,y}^k\}$ with elements represent the depth of pixel $(x,y)$ for the $k$-th part. 
The resulting translation $T_k$ for the $k$-th part is:
\begin{equation}
\label{eqa:transretrive}
T_k = \pi^{-1}\left(\sum_{(x,y) \in G}[x \cdot p_{x,y}^k, y \cdot p_{x,y}^k, d_{x,y}^k \cdot p_{x,y}^k]\right)
\end{equation}
During testing, the produced 3D model is our output.
During training, rather than employing 3D supervision, we use a differentiable renderer to transform 3D representations into images. We render our 3D representation, compare the rendered result, $R$, with the input image, $I$, and then back-propagate through the renderer. The objective we use here is mean squared error pixel reconstruction loss:
\begin{equation}
\label{eqa:reconloss}
\mathcal{L}_r = \frac{1}{|G|}\sum_{(x,y)\in G}(I_{x,y} - R_{x,y})^2
\end{equation}
\section{Consistency Constraints}
\label{sec:consistency}

\begin{figure}[t]
  \centering
    \begin{subfigure}[t]{1.\linewidth}
      \centering
      \includegraphics[width=0.95\linewidth]{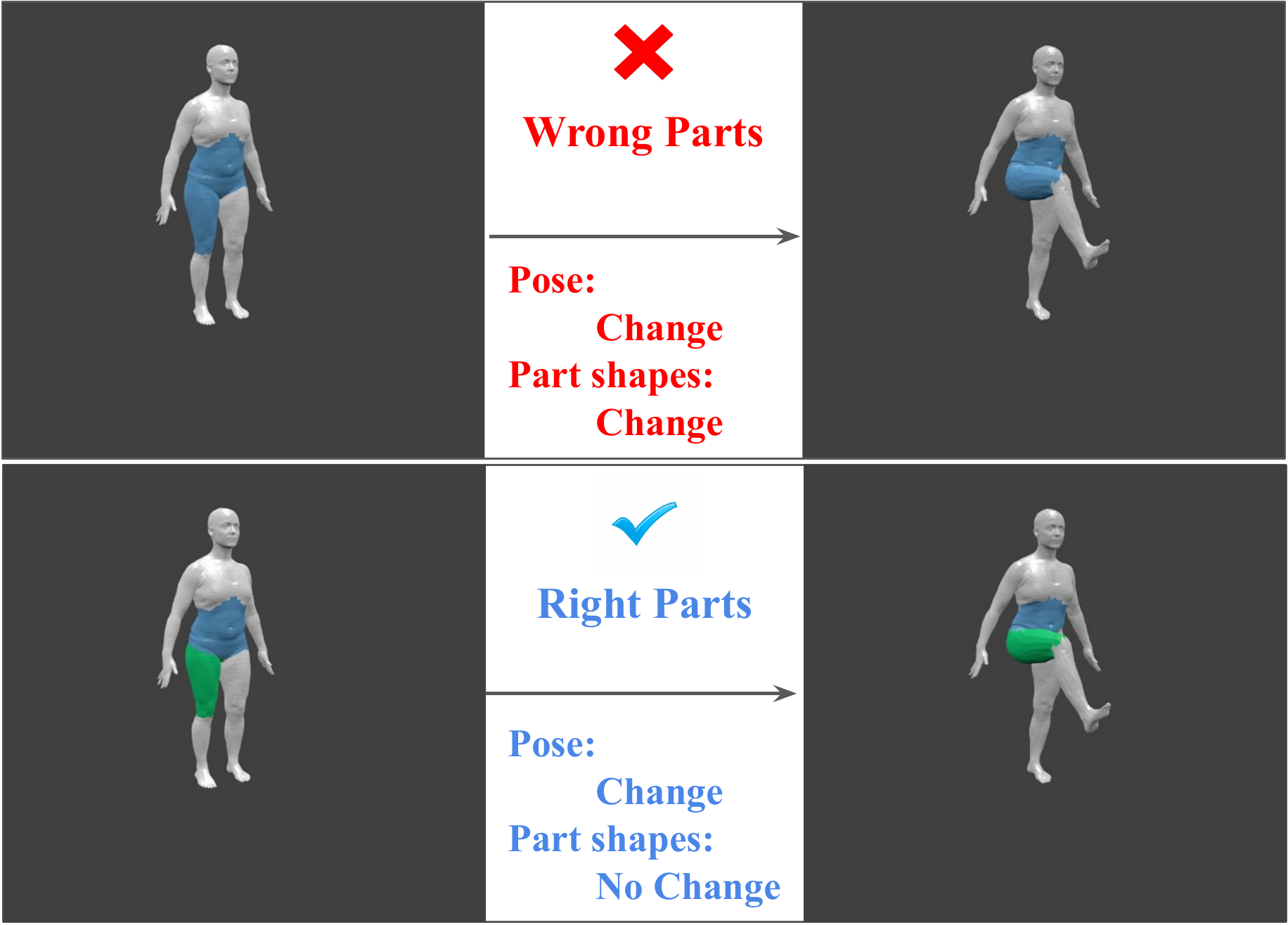}
      \caption{Pose Consistency}
      \label{fig:posecons}
    \end{subfigure}
    \begin{subfigure}[t]{1.\linewidth}
      \centering
      \includegraphics[width=0.95\linewidth]{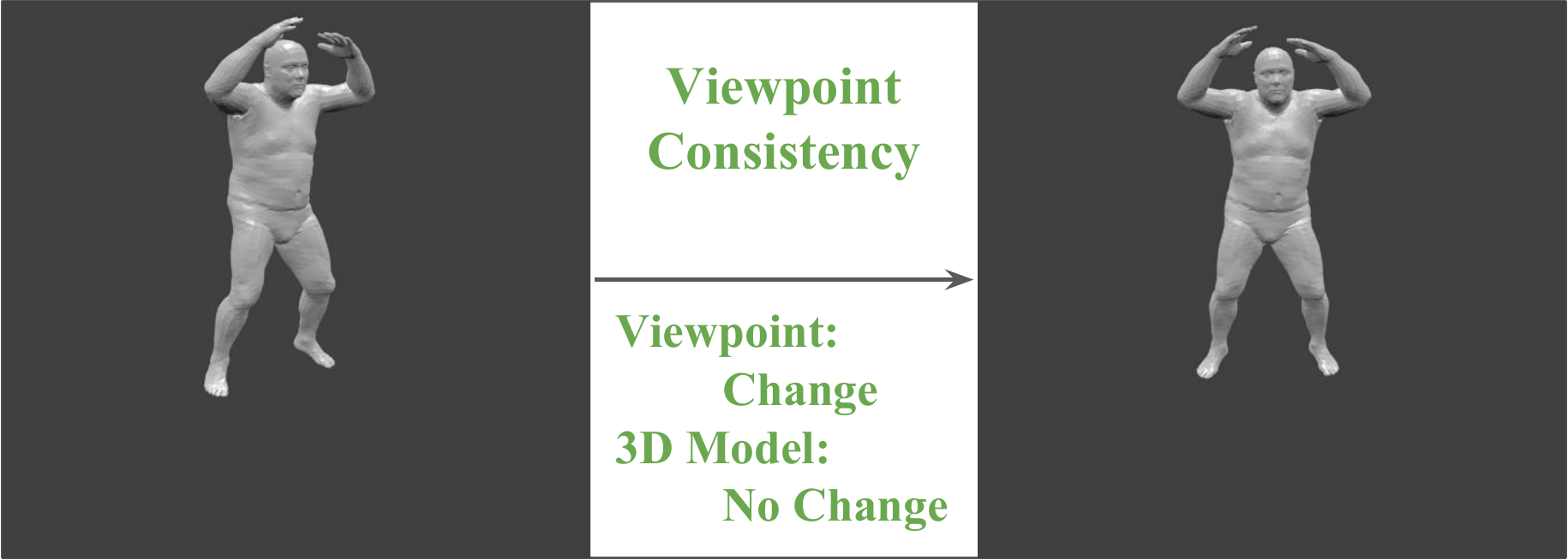}
      \caption{Viewpoint Consistency}
      \label{fig:vpcons}
    \end{subfigure}
  \caption{Consistency constraints enforced in our architecture. \subref{fig:posecons}: Two ways of segmenting a portion of the human body into parts. The segmentation in the bottom row is preferable because the shapes of parts remain unchanged when the pose of the person changes. \subref{fig:vpcons}: Images of a single individual from $2$ viewpoints. Although the rendered images are clearly different, the retrieved 3D model should be the same.}
\end{figure}

\subsection{Pose Consistency}
\label{sec:posecons}
Although we learn part-based models to reconstruct 3D objects, we do not use any part supervision or keypoint annotations during training.
Instead, we reflect on the way humans split an articulated body into multiple parts.
In Figure~\ref{fig:posecons}, we show $2$ different ways to split a portion of the human body into parts. The strategy shown in the bottom row produces semantic parts (abdomen and thigh) whereas the strategy shown in the top row does not. A critical characteristic of the strategy shown in the bottom row is that, when the person's pose changes, the shape of each part remains almost the same. 
This gives us an essential hint on how to split semantic parts without supervision: For a pair of images of the same person in $2$ poses, the $2$ predicted sets of parts should have the same shape.
In practice, we use a pair of images from the same viewpoint containing the same object in $2$ different poses for training. We argue that collecting this kind of supervision is trivial, since we can simply use $2$ frames from a video of a moving object filmed by a static camera.

\subsection{Viewpoint Consistency}
\label{sec:vpcons}
Learning 3D shape from a single image is an ill-posed problem. There exist an infinite number of possible 3D models that yield the same 2D projection.
For the sake of learning correct shapes, we need our predicted 3D models to be consistent across viewpoints during training.
Specifically, we use a pair of images from $2$ different viewpoints for the same object (in the same pose) during training. The goal is to predict the same 3D model from these $2$ viewpoints (as shown in Figure~\ref{fig:vpcons}).
Previous works have investigated this consistency and used cycle construction~\cite{kato2018neural, yan2016perspective} or a loss term~\cite{suwajanakorn2018discovery} to implement the constraint.

\subsection{Constraint Implementation}
\label{sec:consimp}
Combining the above constraints, each training example consists of a quadruplet of images, comprising the same object in $2$ poses seen from $2$ viewpoints.
We index the two poses by $a$ and $b$ and the two viewpoints by $0$ and $1$.
Given a quadruplet during training, Cerberus will output
$4$ shape latents (referred to as $S^{a0}$, $S^{a1}$, $S^{b0}$, and $S^{b1}$)

To enforce the pose consistency constraint, we randomly select elements from the $4$ shape latents corresponding to the $4$ images in the training quadruplet to form the shape latent $\tilde S$ used for rendering. We can formulate $\tilde S$ as:
\begin{equation}
\label{eqa:shapelatent}
  \tilde S = \sum_{x \in Q} \mathbbm{1}_Z(x) \cdot S^{x} 
\end{equation}
where $Q=(a0,a1,b0,b1)$ is the quadruplet and $Z$ is a vector whose elements are sampled from a uniform categorical distribution with $Q$ as categories.

The viewpoint constraint implementation consists of $2$ components. The first component follows the common design of rendering the same 3D model from $2$ viewpoints and comparing them with ground truth images from these viewpoints. Because our model predicts translation parameters, we make a slight change to this design. We render the same rotated mesh with different translations for different viewpoints. Thus, the reconstruction loss for the input in pose $a$ from viewpoint $0$ is:
\begin{equation}
    \mathcal{L}_r^{a0} = \frac{1}{2}(\mathcal{L}_r^{a0 \rightarrow a0} + \mathcal{L}_r^{a0 \rightarrow a1})
\end{equation}
where $\mathcal{L}_r^{a0 \rightarrow a1}$ stands for the reconstruction loss of rendering the rotated mesh produced from image $a0$ from viewpoint $1$ with translation predicted from input $a1$. We use different translations for different viewpoints because the model may predict incorrect translations at the early stages of training. Using the translation predicted from a single viewpoint only guarantees that the object is visible from this viewpoint. From a different viewpoint, the renderer may not see the object. In this case, we would be unable to get a gradient from the renderer, so training would collapse.

The second component encourages the translations to be consistent across viewpoint, using a mean squared error loss to penalize inconsistent translations.
For each pair of viewpoints, the translation loss term is:
\begin{equation}
\label{eqa:transloss}
  \mathcal{L}_t = \frac{1}{N}\sum_i^N(T_i^0 - T_i^1)^2
\end{equation}
where $T_i^0$ and $T_i^1$ are predicted translations of the $i$-th part from $2$ viewpoints. Since we have $2$ poses, the full translation loss term for each quadruplet is:
\begin{equation}
\label{eqa:translosstotal}
  \mathcal{L}_t^\prime = \frac{1}{2}(\mathcal{L}_t^{a} + \mathcal{L}_t^{b})
\end{equation}
where $\mathcal{L}_t^{a}$ represents $\mathcal{L}_t$ for pose $a$ and $\mathcal{L}_t^{b}$ represents $\mathcal{L}_t$ for pose $b$.
The total reconstruction loss is:
\begin{equation}
\label{eqa:totalrecon}
  \mathcal{L}_r^\prime = \frac{1}{4}(\mathcal{L}_r^{a0} + \mathcal{L}_r^{a1} + \mathcal{L}_r^{b0} + \mathcal{L}_r^{b1})
\end{equation} 
\section{Experiments}
\label{sec:exp}

\begin{table*}[t]
    \centering
    \begin{tabular}{c|ccc|ccc}
        \toprule
        \textbf{Input} & \textbf{NMR} & \textbf{NMRs} & \textbf{NMRr} & \textbf{Ours} & \textbf{Parts} & \textbf{Turn} \\
        \midrule
        \includegraphics[width=0.12\linewidth]{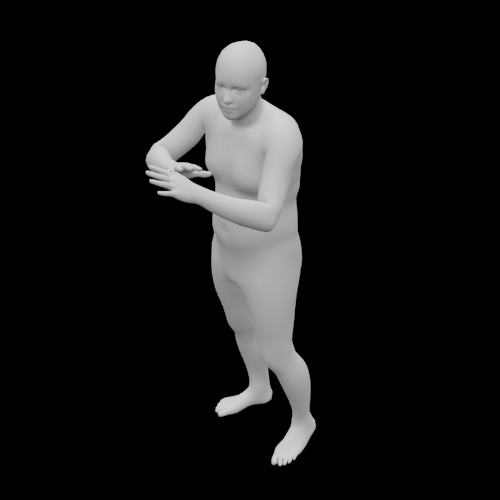} &
            \includegraphics[width=0.12\linewidth]{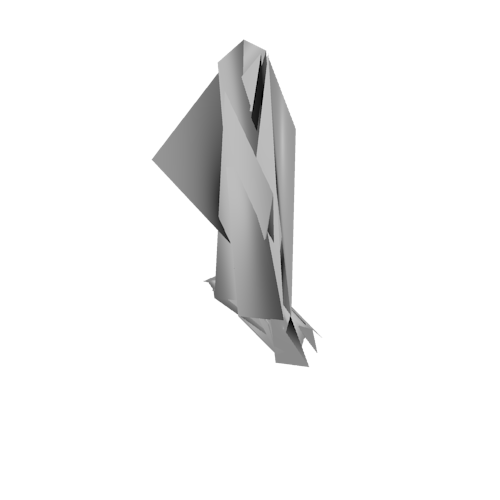} &
            \includegraphics[width=0.12\linewidth]{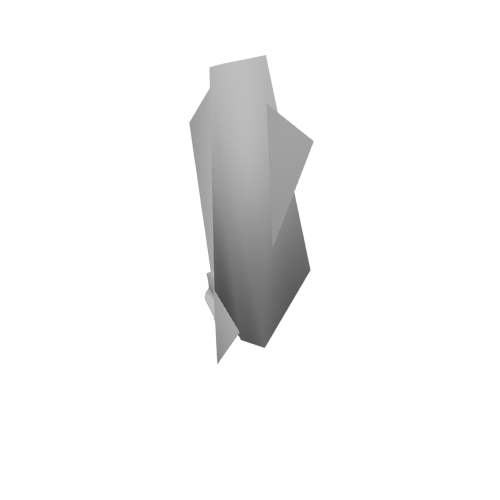} &
            \includegraphics[width=0.12\linewidth]{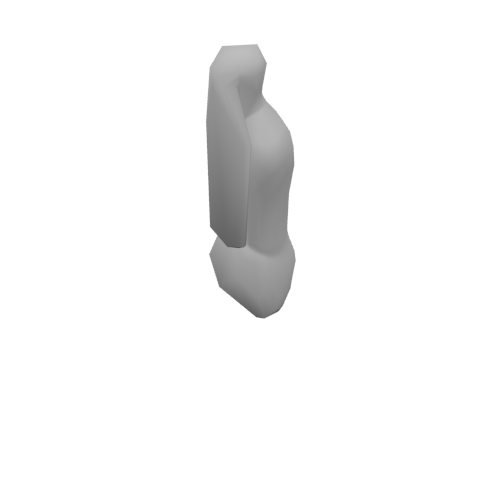} &
            \includegraphics[width=0.12\linewidth]{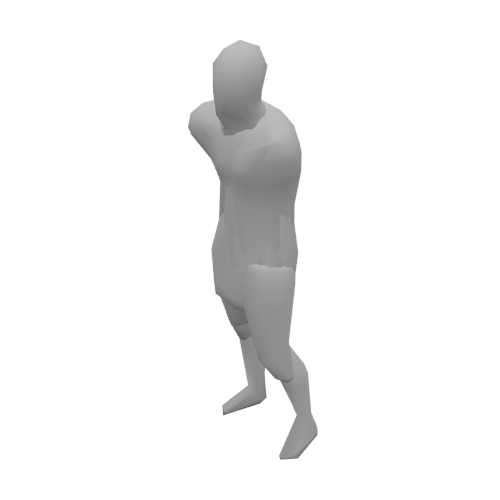} &
            \includegraphics[width=0.12\linewidth]{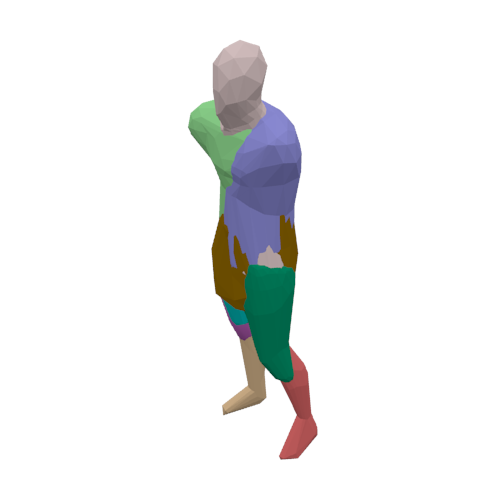} &
            \includegraphics[width=0.12\linewidth]{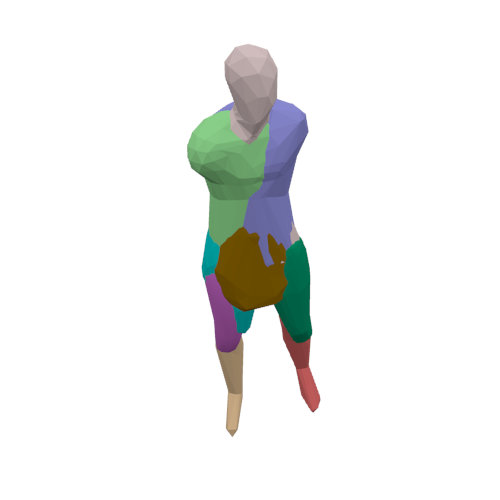} \\
        \includegraphics[width=0.12\linewidth]{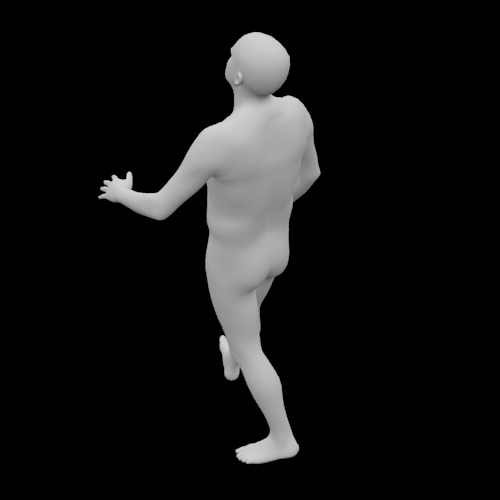} &
            \includegraphics[width=0.12\linewidth]{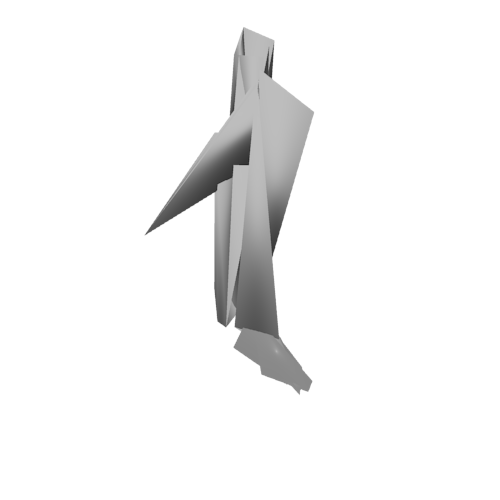} &
            \includegraphics[width=0.12\linewidth]{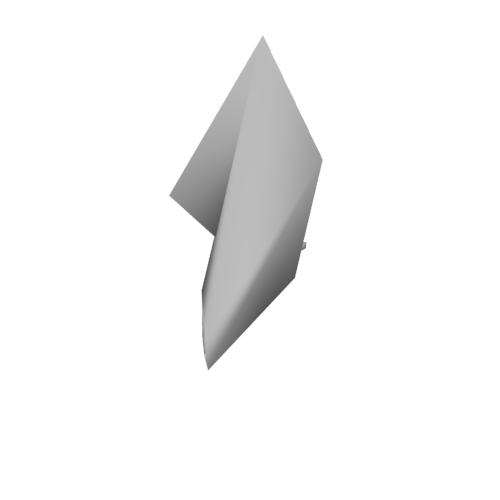} &
            \includegraphics[width=0.12\linewidth]{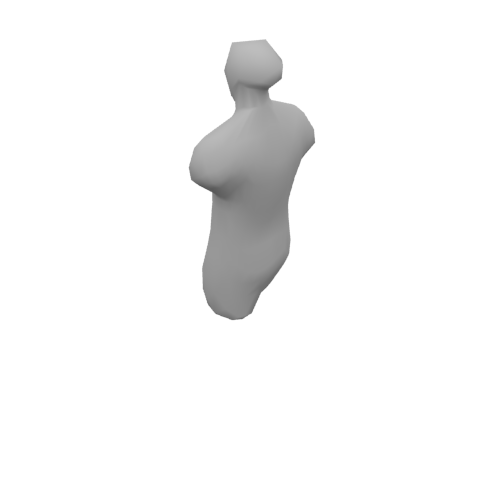} &
            \includegraphics[width=0.12\linewidth]{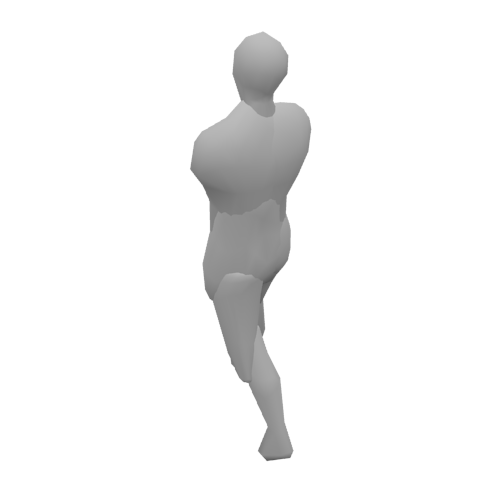} &
            \includegraphics[width=0.12\linewidth]{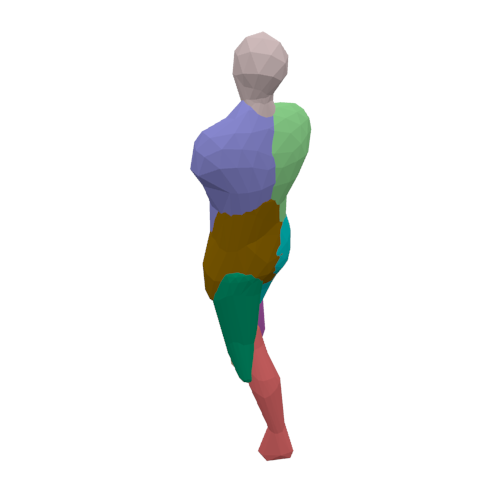} &
            \includegraphics[width=0.12\linewidth]{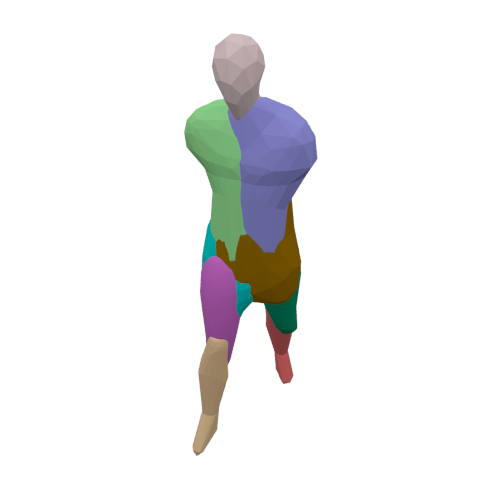} \\
        \includegraphics[width=0.12\linewidth]{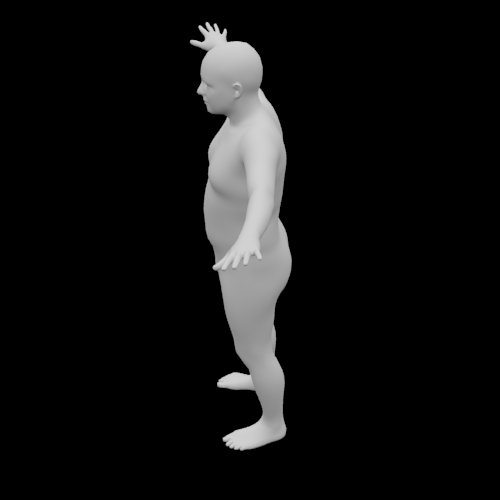} &
            \includegraphics[width=0.12\linewidth]{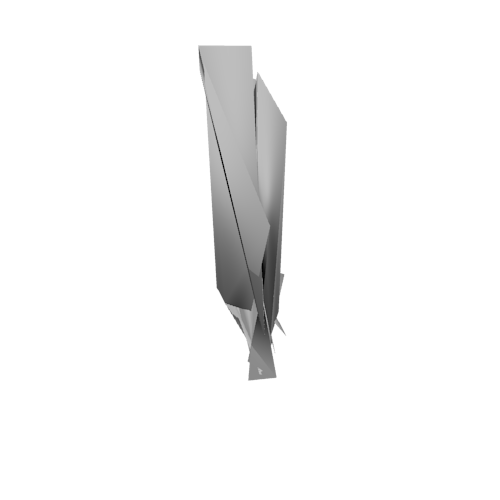} &
            \includegraphics[width=0.12\linewidth]{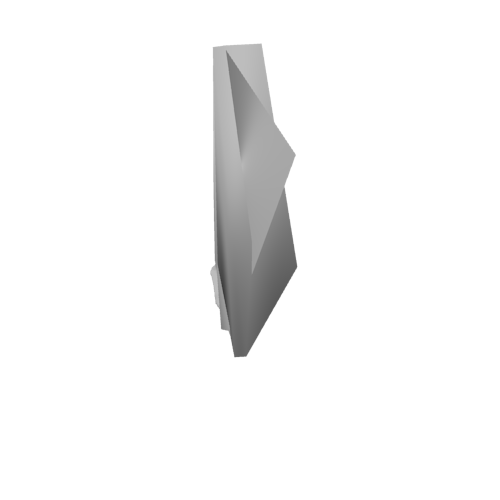} &
            \includegraphics[width=0.12\linewidth]{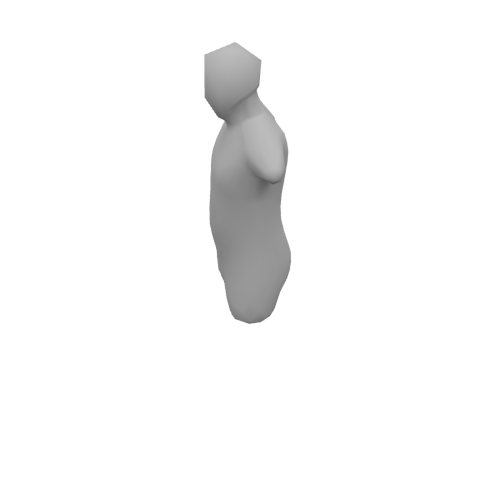} &
            \includegraphics[width=0.12\linewidth]{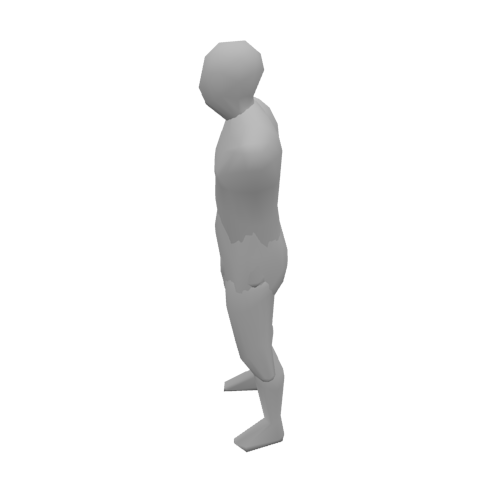} &
            \includegraphics[width=0.12\linewidth]{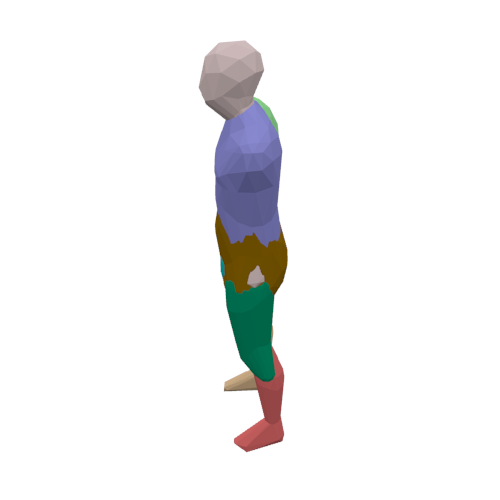} &
            \includegraphics[width=0.12\linewidth]{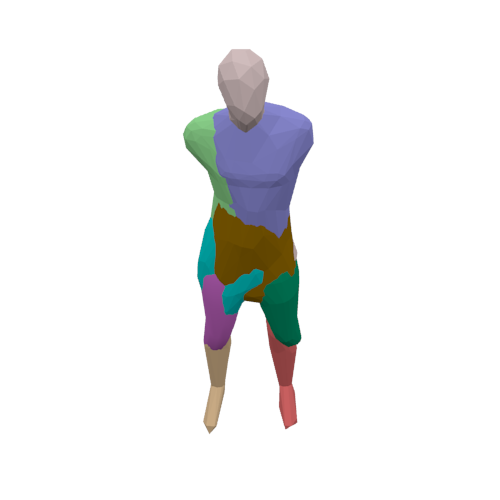} \\
        \includegraphics[width=0.12\linewidth]{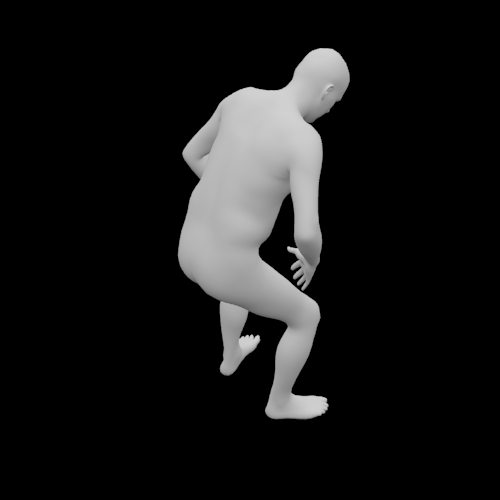} &
            \includegraphics[width=0.12\linewidth]{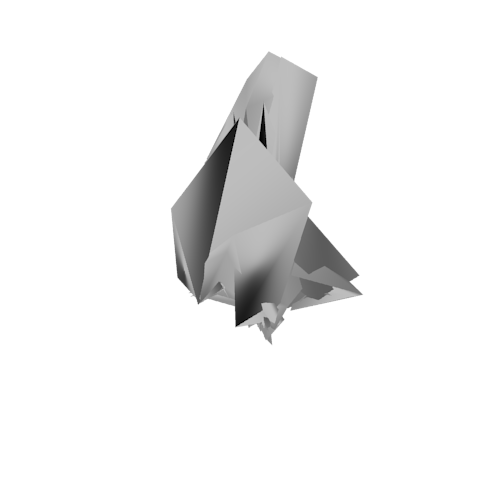} &
            \includegraphics[width=0.12\linewidth]{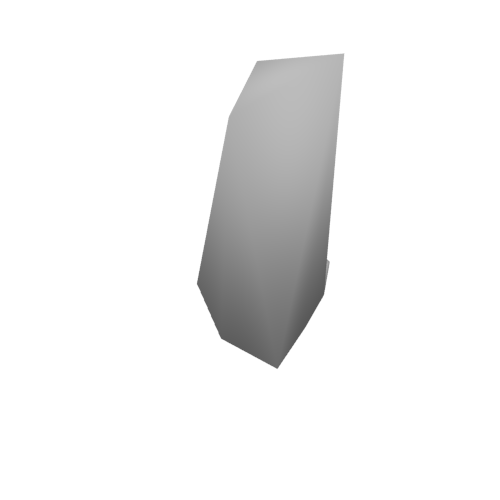} &
            \includegraphics[width=0.12\linewidth]{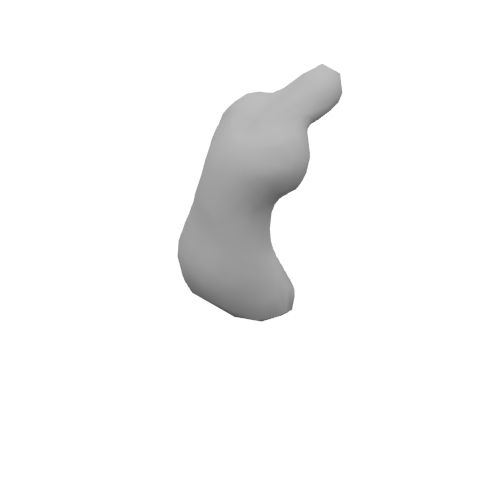} &
            \includegraphics[width=0.12\linewidth]{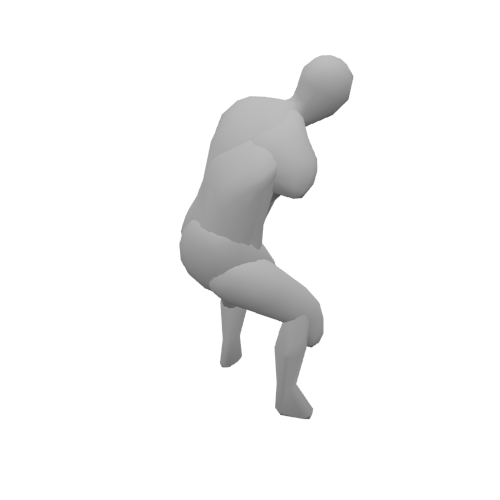} &
            \includegraphics[width=0.12\linewidth]{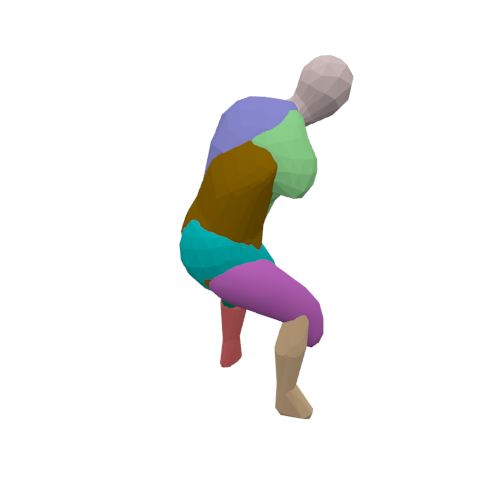} &
            \includegraphics[width=0.12\linewidth]{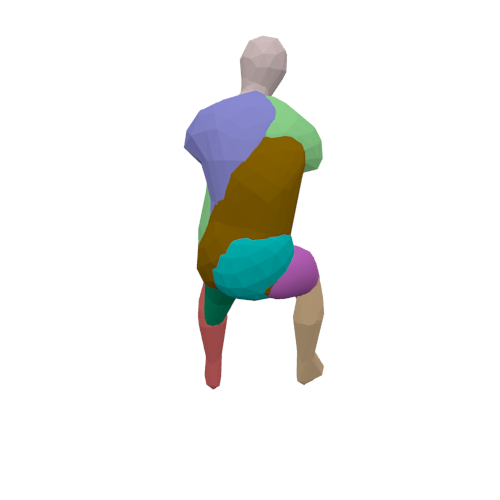} \\
        \includegraphics[width=0.12\linewidth]{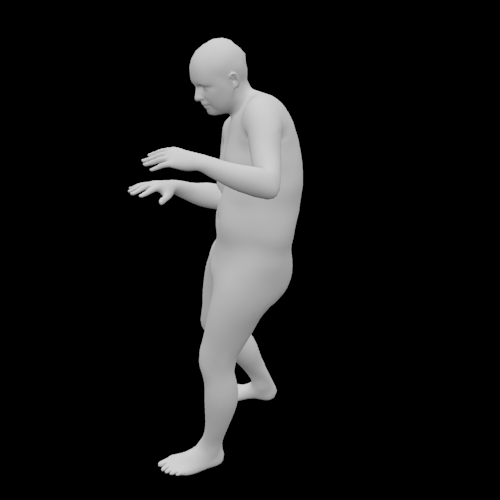} &
            \includegraphics[width=0.12\linewidth]{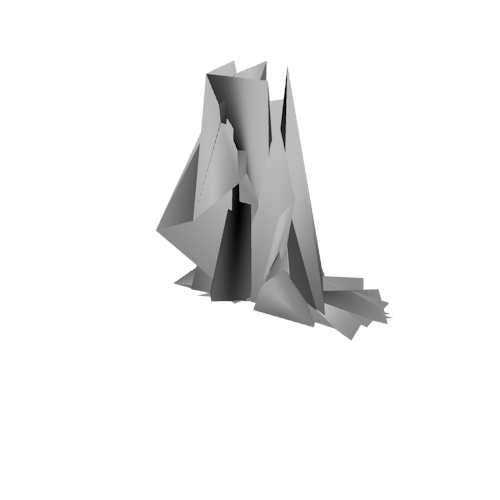} &
            \includegraphics[width=0.12\linewidth]{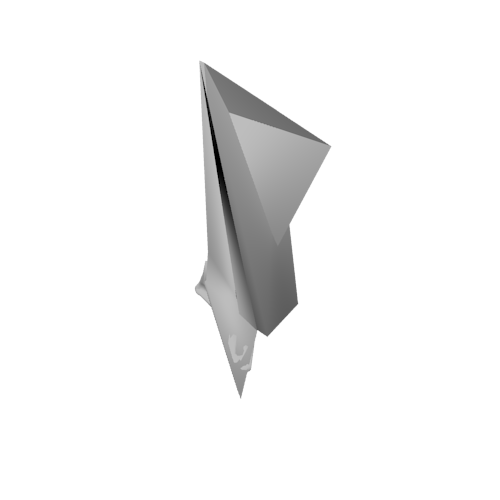} &
            \includegraphics[width=0.12\linewidth]{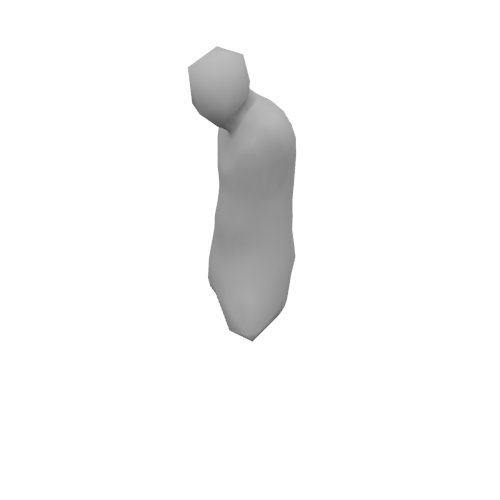} &
            \includegraphics[width=0.12\linewidth]{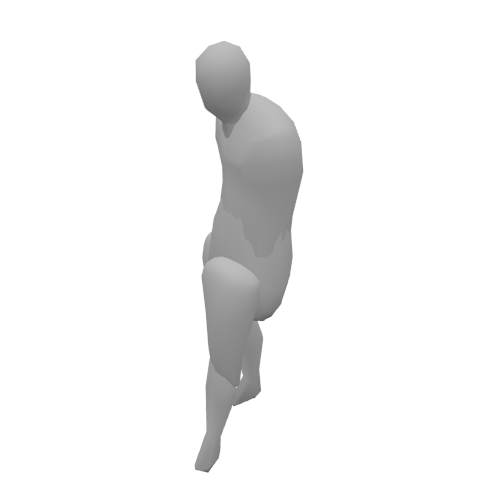} &
            \includegraphics[width=0.12\linewidth]{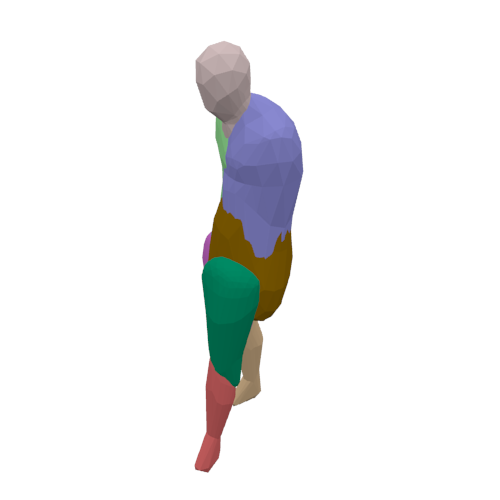} &
            \includegraphics[width=0.12\linewidth]{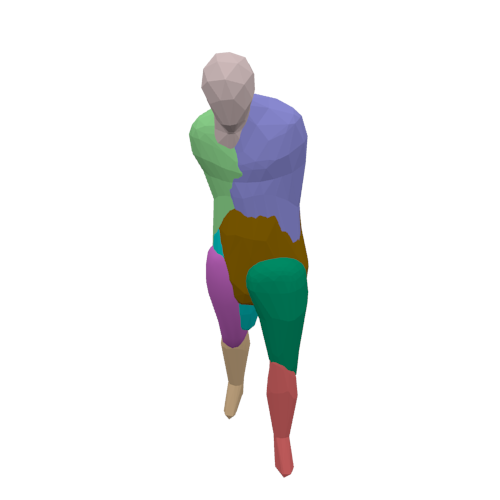} \\
        \bottomrule
    \end{tabular}
    \caption{3D Human Model Reconstructions. On the left are input images. In the middle are baseline results.
    NMR is the architecture and implementation from~\cite{kato2018neural} without a smoothness loss. NMRs adds smoothness loss to NMR.
    NMRr is our re-implementation of NMR using the same renderer and images with shading as used by Cerberus.
    We also visualize our parts in different colors in the middle column of the right group. The Turn column shows Cerberus's outputs rendered from a different viewpoint.}
    \label{tab:humanrecon}
\end{table*}

\paragraph{Human Dataset.} This dataset contains 3D human models in diverse body poses. We use SMPL~\cite{loper2015smpl}, a parameterized deformable human model, to generate all the example meshes.
In particular, the parameters of SMPL are fit to frames of human action video clips from Human3.6M~\cite{ionescu2014human3} using Mosh~\cite{loper2014mosh}. In our experiments, we use the fitted results from~\cite{hmrKanazawa17}.
We split the data into train and test splits. 
The train split comprises $19{,}500$ pairs of body poses with $5$ subjects (S1, S5, S6, S7, S8 in Human3.6M). Each pose is rendered from $2$ different viewpoints. We fix the elevation angle and distance to the origin of all viewpoints but vary the azimuth angles. On scene set-ups, we use a directional light following the direction of the camera and an ambient light, all in white.
In the test split, we use $2$ unseen subjects (S9, S11 in Human3.6M). We render $810$ different poses in total, each from $4$ viewpoints.

\paragraph{Animal Dataset.} This dataset consists of 3D models of quadrupeds. Compared with the Human dataset, it has more variance in shape but less variance in pose. Each example is generated by a deformable model, SMAL~\cite{zuffi20173d}, for quadrupeds.
We use $41$ different animals released by~\cite{zuffi20173d}. Each of them has $47$ poses in the train split and $3$ poses in the test split. We render each training pose from $4$ viewpoints and each test pose from $8$ viewpoints. In total, the dataset contains $38{,}540$ training quadruplets (as used in Section~\ref{sec:consimp}) and $984$ test examples. The scene set-ups are the same as the human dataset.

\paragraph{Implementation Details.} Our down-sampling network is ResNet-10-v1. The number of channels are $64$, $128$, $256$, $512$ for different feature map resolutions. The up-sampling network has $3$ transposed convolution layers with skip connections from down-sampling layers of the same resolution. We use a spherical mesh with $162$ vertices and $320$ triangles as the starting point of the deformation for each part. We set the number of parts to $9$ for all experiments.
During training, we use $2$ additional regularization loss terms.
One is a background loss
$\mathcal{L}_b = \frac{1}{N} \sum_k\sum_{x,y}p_{x,y}^k b_{x,y}$, where $b_{x,y}$ indicates whether pixel $(x,y)$ is in the background. This loss helps avoid situations where a part is out of the renderer's scope and no gradient will pass through.
The other is a smoothness loss $\mathcal{L}_s = \sum_{\theta_i \in \epsilon} (\cos \theta_i + 1)^2$ following~\cite{kato2018neural}. $\epsilon$ is the set of dihedral angles of the output mesh. This loss helps encourage neighboring vertices to have similar displacement.
We use Adam with learning rate $0.0005$ and batch size $16$ to optimize the weighted sum of all the loss terms, with weights $(\lambda_r, \lambda_t, \lambda_b, \lambda_s)$ set to $(1, 1, 1, 0.0001)$ for all experiments. We train our models for $100{,}000$ steps. Unlike~\cite{kato2018neural}, we use a differentiable renderer based on gradients of barycentric coordinates~\cite{Genova_2018_CVPR}.

\subsection{3D Human Reconstruction}
\label{sec:humanrecon}

We first test our model on single image 3D human reconstruction.
This task is a standard benchmark for 3D visual perception methods. The goal of this task is to extract 3D models of the object in the given image.
We measure the quality of the predicted 3D models by voxel IoU (intersection-over-union), following~\cite{yan2016perspective}. Since our 3D models are meshes, we transform the predicted meshes and the ground-truth meshes into $32\times32\times32$ voxel grids. Because Cerberus predicts multiple parts, we take the union of all the part voxels as the output voxel.

We provide baseline results with the Neural Mesh 3D Renderer (NMR)~\cite{kato2018neural}. Because we found that the smoothness loss proposed in~\cite{kato2018neural} performs poorly for curved surfaces, we train NMR models both with and without this smoothness loss. We also recognize that NMR uses only silhouette supervision, whereas the differentiable renderer we use for Cerberus provides gradients for shaded surfaces. To exclude confounds related to the choice of renderer, we also re-implement NMR using our renderer on images rendered with shading. We visualize some example 3D outputs of all the baselines and Cerberus on the test split in Table~\ref{tab:humanrecon}.

As shown in Table~\ref{tab:humanrecon}, Cerberus predicts smooth 3D meshes that are visually more similar to the human in the input.
Compared with NMR, which mainly reconstructs the outline shape of the torso, Cerberus can produces more details of the body, \eg the legs. We hypothesize that this improvement is related to the flexibility of part-based modeling.
More importantly, we find that, although our model is trained without any part annotations, it can predict semantic parts of the human body. For instance, the beige part shown in Table~\ref{tab:humanrecon} is clearly recognizable as the head.
Similarly, we find parts representing legs in Cerberus's outputs. Cerberus can even separate lower legs and thighs into different parts. We believe that this segmentation arises from our pose consistency constraint.
We have both examples with bent knees and examples with straight legs in our training split. In order to model the body accurately, Cerberus must learn to separate these two parts. We also notice that Cerberus produces plausible results for invisible parts (as shown in the third row of Table~\ref{tab:humanrecon}), suggesting that the neural network implicitly incorporates the prior of human's body shape so that it outputs two legs even when only one leg is visible.

\begin{table}[t]
    \centering
    \begin{tabular}{c|cc|c}
        \toprule
        \textbf{Model} & \textbf{Human} & \textbf{Hard Human} & \textbf{Animal} \\
        \midrule \midrule
        NMR & $0.2596$ & - & $0.3000$ \\
        NMRs & $0.2233$ & - & $0.2574$ \\
        NMRr & $0.3084$ & - & $0.3201$ \\
        \midrule
        Cerberus & $0.4970$ & \boldmath $0.4728$ & \boldmath $0.4255$ \\
        Free Cerberus & \boldmath $0.5099$ & $0.4365$ & $0.4196$ \\
        \bottomrule
    \end{tabular}
    \caption{Single image 3D reconstruction test results on $2$ datasets. We use voxel IoU (intersection-over-union) as our metric. Higher is better. NMR refers the model proposed by~\cite{kato2018neural}. NMRs is NMR with smoothness loss. NMRr is our re-implementation using the same renderer as Cerberus. Free Cerberus is Cerberus trained without pose consistency. Hard Human reflects accuracy when reconstructing all poses in the test set using the same set of parts. Hard Human results are shown only for Cerberus, since NMR does not model individual parts.}
    \label{tab:voxiou}
\end{table}

Quantitatively, Cerberus predicts 3D meshes with greater similarity to the target meshes than previous approaches (Table~\ref{tab:voxiou}). 
Compared with the original NMR, Cerberus achieves double the test IoU. Our re-implementation of NMR has a higher accuracy than the original NMR, suggesting that reconstructing shaded images helps learn better shapes. Nonetheless, Cerberus outperforms this re-implemented NMR by a substantial margin.

\subsection{Transferable Parts Among Poses}
\label{sec:transparts}

\begin{figure}[t]
    \centering
    \begin{subfigure}[t]{.95\linewidth}
        \centering
        \includegraphics[width=0.3\linewidth]{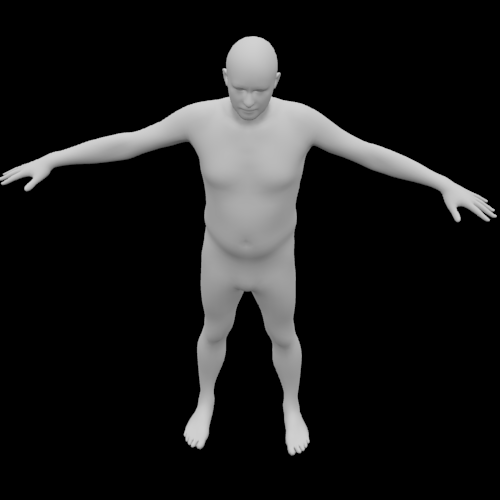}
        \hspace{0.01\linewidth}
        \includegraphics[width=0.3\linewidth]{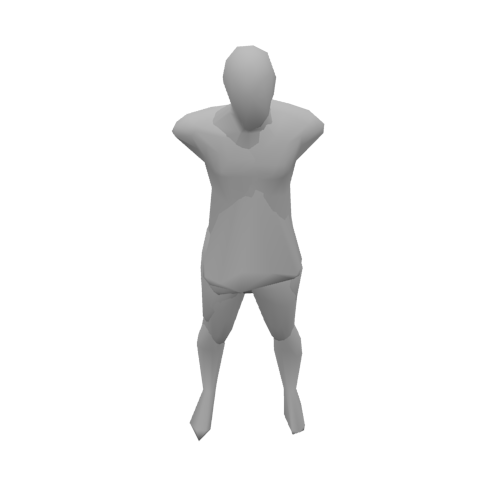}
        \hspace{0.01\linewidth}
        \includegraphics[width=0.3\linewidth]{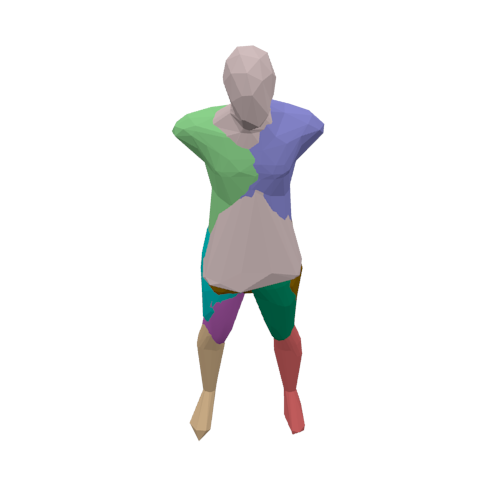}
        \caption{Canonical Inputs and Reconstruction}
        \label{fig:canonpose}
    \end{subfigure}
    \begin{subfigure}[t]{.95\linewidth}
        \centering
        \includegraphics[width=0.3\linewidth]{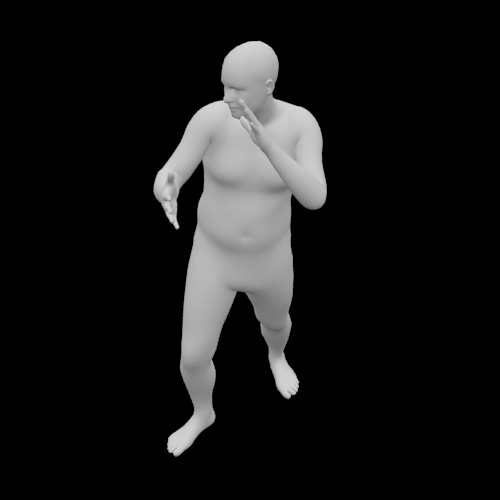}
        \hspace{0.01\linewidth}
        \includegraphics[width=0.3\linewidth]{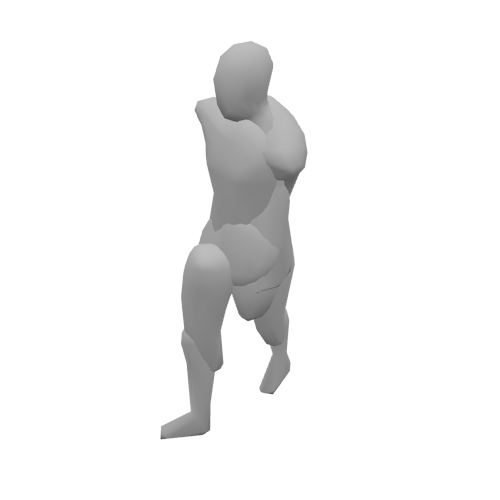}
        \hspace{0.01\linewidth}
        \includegraphics[width=0.3\linewidth]{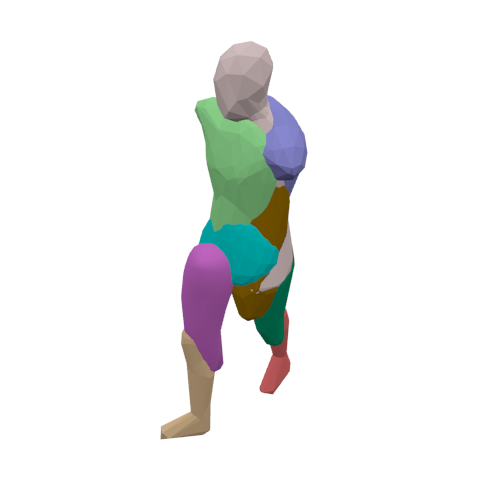}
    \end{subfigure}
    \begin{subfigure}[t]{.95\linewidth}
        \centering
        \includegraphics[width=0.3\linewidth]{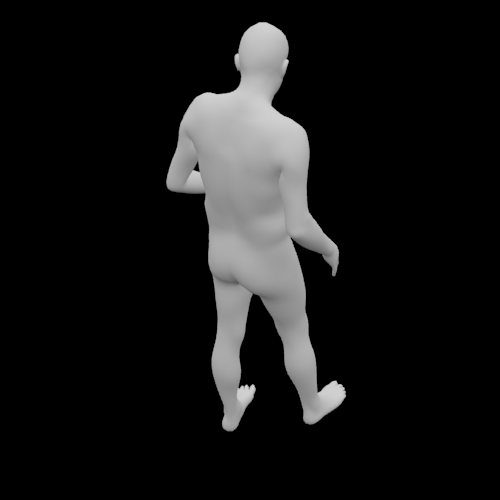}
        \hspace{0.01\linewidth}
        \includegraphics[width=0.3\linewidth]{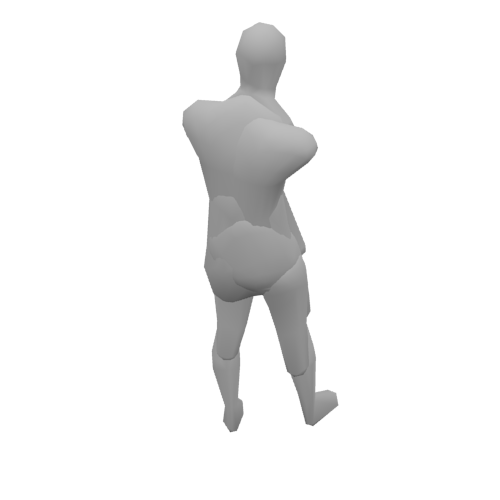}
        \hspace{0.01\linewidth}
        \includegraphics[width=0.3\linewidth]{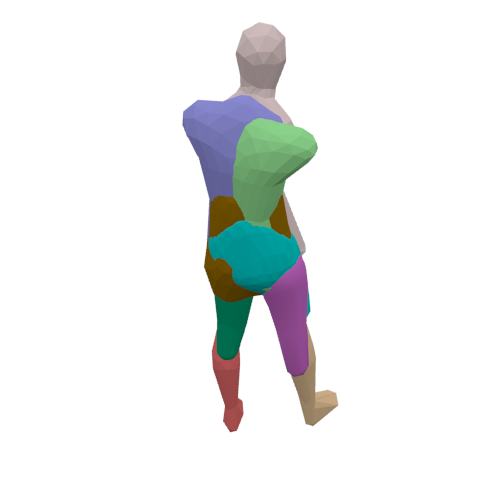}
    \end{subfigure}
    \begin{subfigure}[t]{.95\linewidth}
        \centering
        \includegraphics[width=0.3\linewidth]{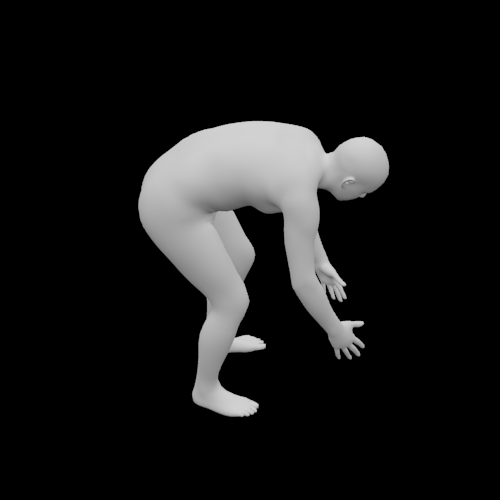}
        \hspace{0.01\linewidth}
        \includegraphics[width=0.3\linewidth]{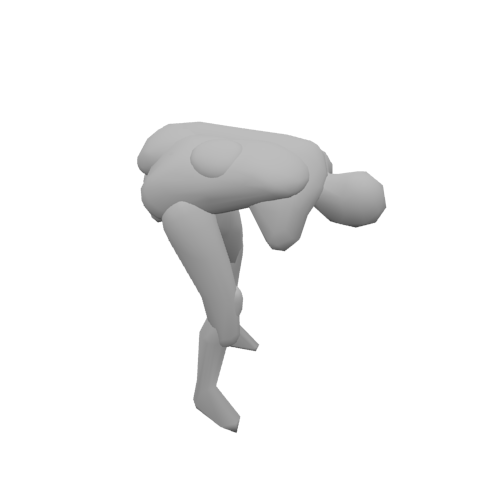}
        \hspace{0.01\linewidth}
        \includegraphics[width=0.3\linewidth]{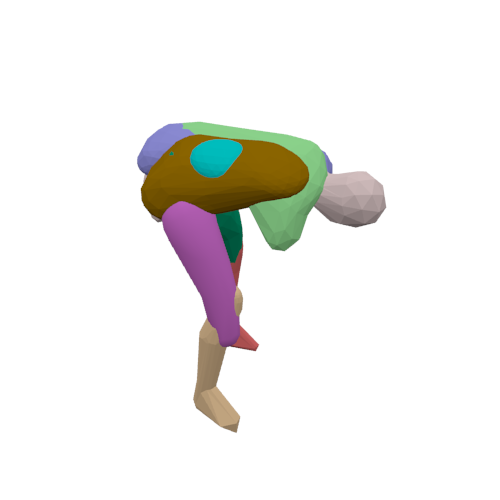}
        \caption{Reconstruction by transformations}
        \label{fig:newpose}
    \end{subfigure}
    \caption{Hard human reconstruction test. We extract a single set of deformed shape meshes from the canonical inputs (\subref{fig:canonpose}) and apply new transformations (rotation and translation) predicted from other images to these meshes to reconstruct new poses (\subref{fig:newpose}). \textbf{Left}: Input images. \textbf{Middle}: Reconstructed 3D outputs. \textbf{Right}: Parts rendered in different colors.}
    \label{fig:humanhard}
\end{figure}

We also devise a benchmark to quantitatively evaluate the accuracy with which Cerberus segments parts.
Instead of computing IoU when reconstructing each test case independently, we perform identity-conditional reconstruction. We first extract the deformed part meshes from $2$ images of the $2$ subjects in the test set. These images contain the canonical pose of the the subjects (shown in Figure~\ref{fig:canonpose}). Then, we reconstruct other examples in the test split by applying predicted rotation and translation to the deformed parts from the canonical pose of the same subject.

The voxel IoU of this more challenging evaluation (``Hard Human") is shown in Table~\ref{tab:voxiou}. Even in this extreme test set-up, Cerberus's accuracy remains high, and our predictions remain better than all baseline methods. This result quantitatively confirms that the pose consistency constraint enforces Cerberus to learn meaningful parts that are transferable among diverse poses. We illustrate the output meshes produced by this evaluation set-up in Figure~\ref{fig:humanhard}.

\begin{figure}
    \centering
    \includegraphics[width=.3\linewidth]{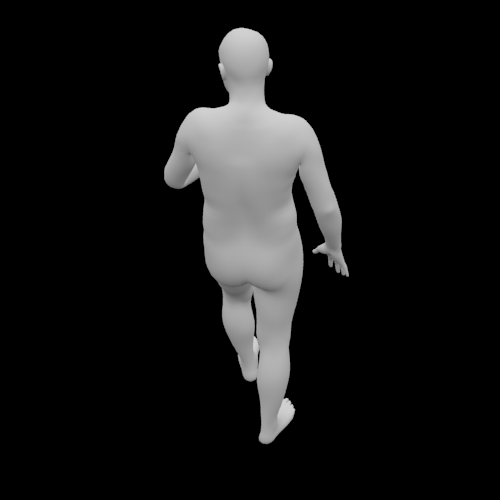}
    \includegraphics[width=.3\linewidth]{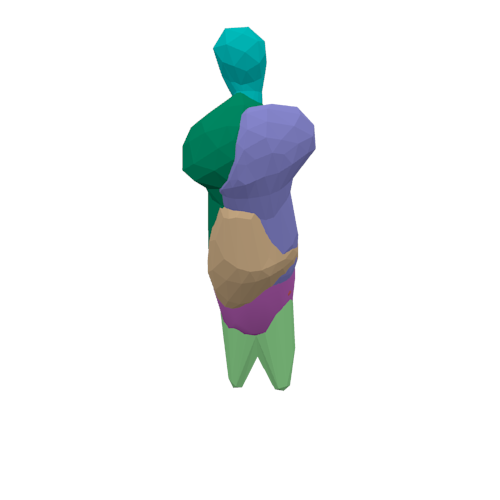}
    \includegraphics[width=.3\linewidth]{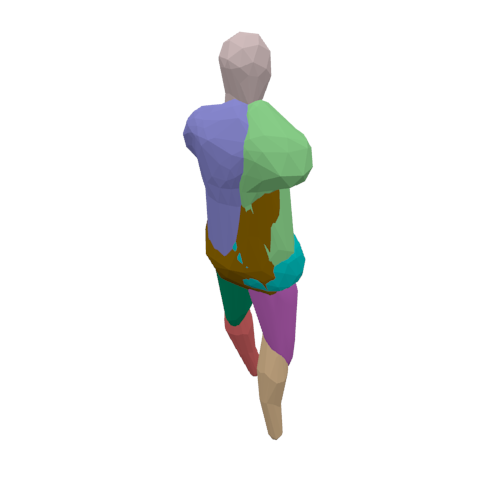}
    \caption{Comparison between Free Cerberus (w/o pose consistency) and Cerberus (w/ pose consistency). \textbf{Left:} Input. \textbf{Middle:} Output of Free Cerberus. \textbf{Right:} Output of Cerberus. Free Cerberus uses a single green part for $2$ legs while Cerberus correctly models $2$ legs.}
    \label{fig:ablation} 
\end{figure}

To validate the effectiveness of pose consistency constraint for learning transferable and semantic parts, we perform an ablation study by training Cerberus without pose consistency constraint. The test IoU of this model (``Free Cerberus") is shown in Table~\ref{tab:voxiou}. Free Cerberus achieves higher accuracy in the standard evaluation than Cerberus with pose consistency. This is unsurprising, given that Free Cerberus has additional freedom in modeling objects. However, in the Hard Human evaluation, Free Cerberus's accuracy is significantly lower than Cerberus, with a large drop relative to the accuracy on standard evaluation. Thus, the pose consistency constraint is important to properly segment parts. We visualize the outputs of Cerberus and Free Cerberus in Figure~\ref{fig:ablation}. We see that Free Cerberus uses a single green part to represent two legs, whereas Cerberus models the legs with separate parts.

\subsection{3D Animal Reconstruction}
\label{sec:animalrecon}
In addition to reconstructing 3D human models from a single image, we also examine Cerberus's ability to reconstruct objects with greater variability in shape.
To this end, we evaluate our method on the animal dataset. This dataset comprises of $41$ different animals, ranging from deer with small heads and long thin limbs to hippopotamuses with large heads and short thick limbs.

We show test IoU on the animal dataset in Table~\ref{tab:voxiou}. Baseline methods perform better on the animal dataset as compared to the human dataset, likely because there is less variability in pose. Nonetheless, Cerberus remains superior. Thus, Cerberus consistently outperforms baselines on objects with either high variability in pose (humans) or high variability in shape (animals).

\begin{figure}[t]
    \centering
    \begin{subfigure}[t]{.95\linewidth}
        \centering
        \includegraphics[width=0.3\linewidth]{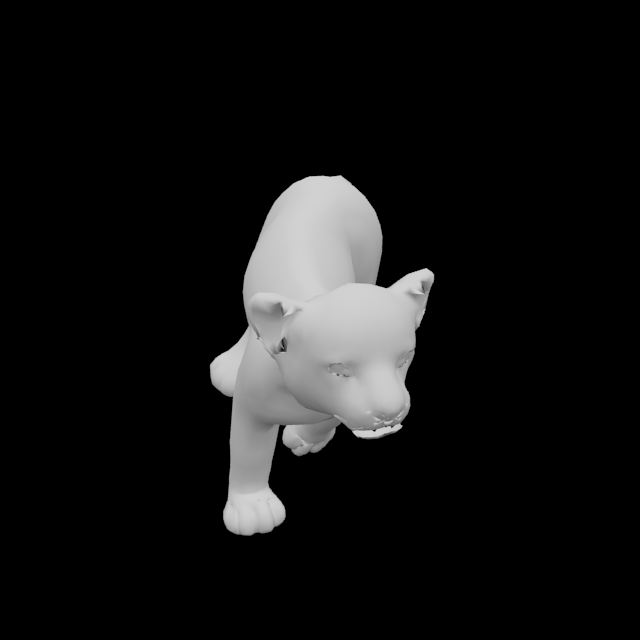}
        \hspace{0.01\linewidth}
        \includegraphics[width=0.3\linewidth]{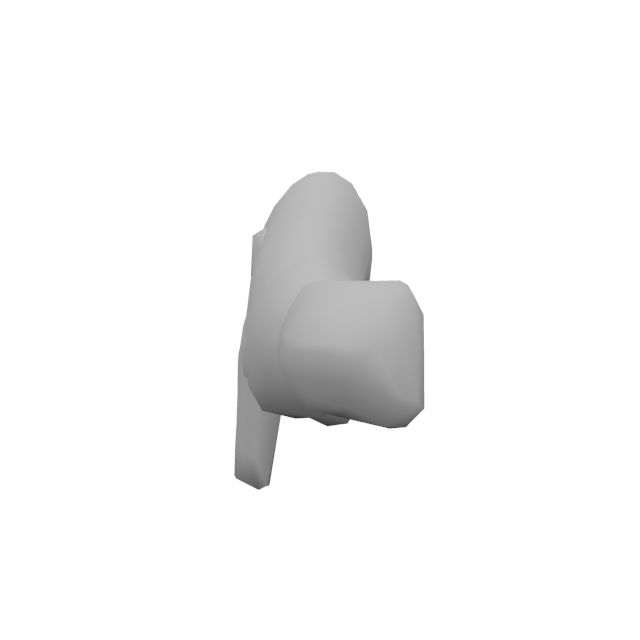}
        \hspace{0.01\linewidth}
        \includegraphics[width=0.3\linewidth]{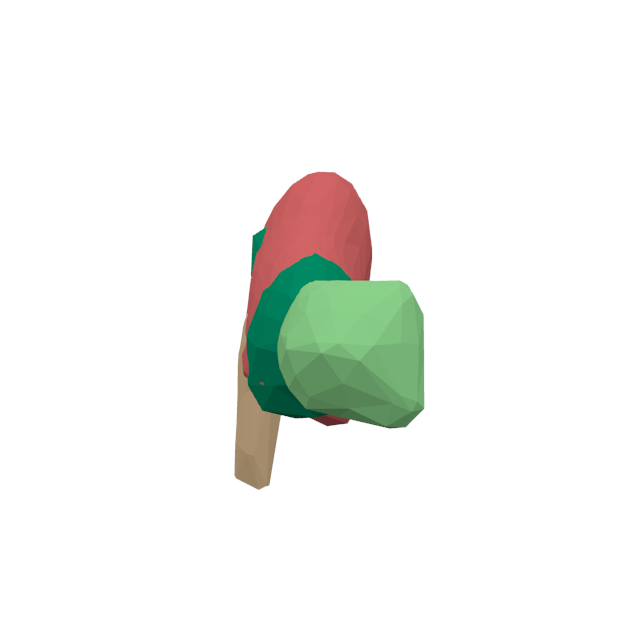}
        \caption{Cougar}
        \label{fig:Cougar}
    \end{subfigure}
    \begin{subfigure}[t]{.95\linewidth}
        \centering
        \includegraphics[width=0.3\linewidth]{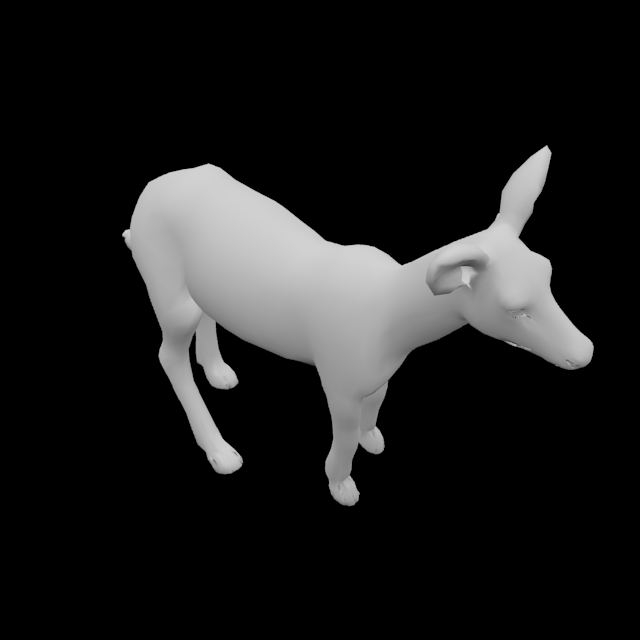}
        \hspace{0.01\linewidth}
        \includegraphics[width=0.3\linewidth]{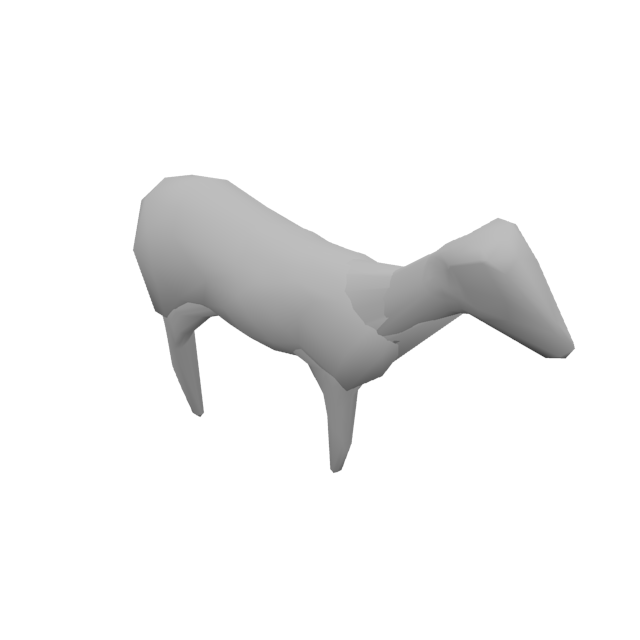}
        \hspace{0.01\linewidth}
        \includegraphics[width=0.3\linewidth]{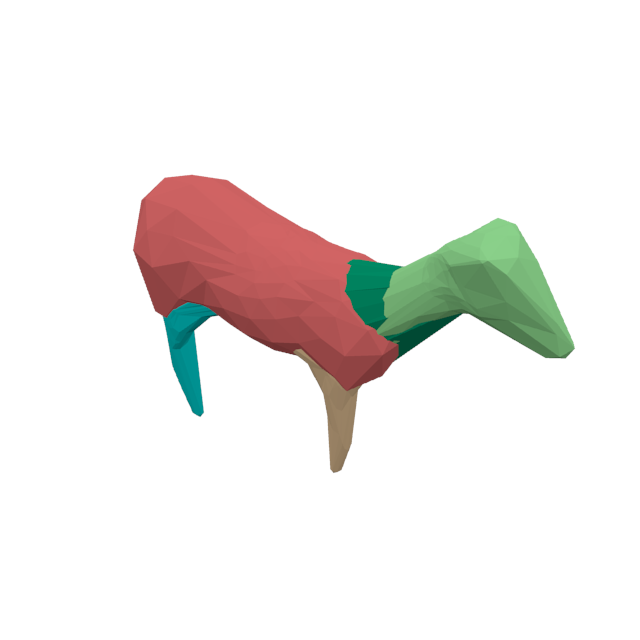}
        \caption{Deer}
        \label{fig:deer}
    \end{subfigure}
    \begin{subfigure}[t]{.95\linewidth}
        \centering
        \includegraphics[width=0.3\linewidth]{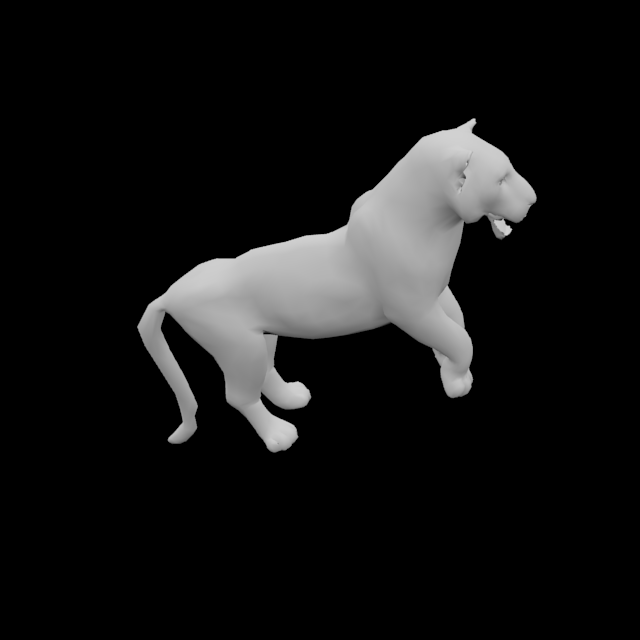}
        \hspace{0.01\linewidth}
        \includegraphics[width=0.3\linewidth]{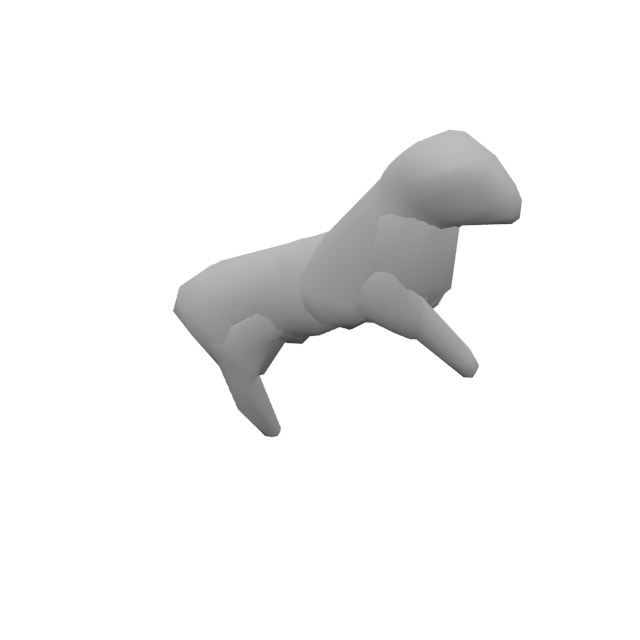}
        \hspace{0.01\linewidth}
        \includegraphics[width=0.3\linewidth]{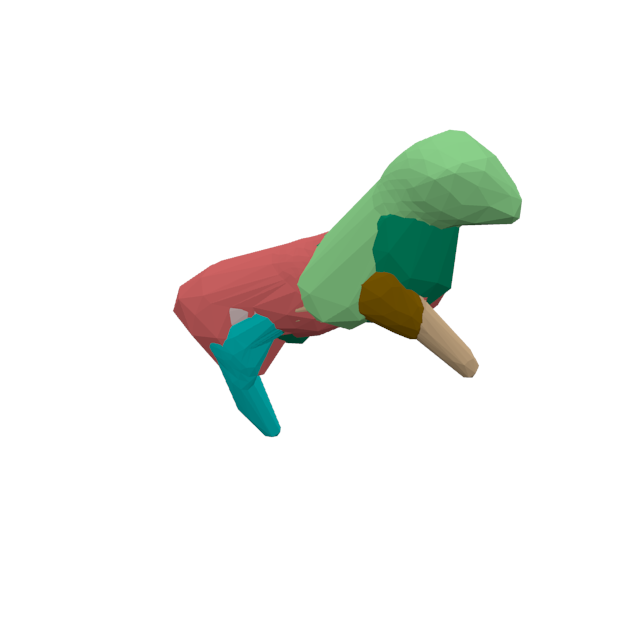}
        \caption{Tiger}
        \label{fig:tiger}
    \end{subfigure}
        \begin{subfigure}[t]{.95\linewidth}
        \centering
        \includegraphics[width=0.3\linewidth]{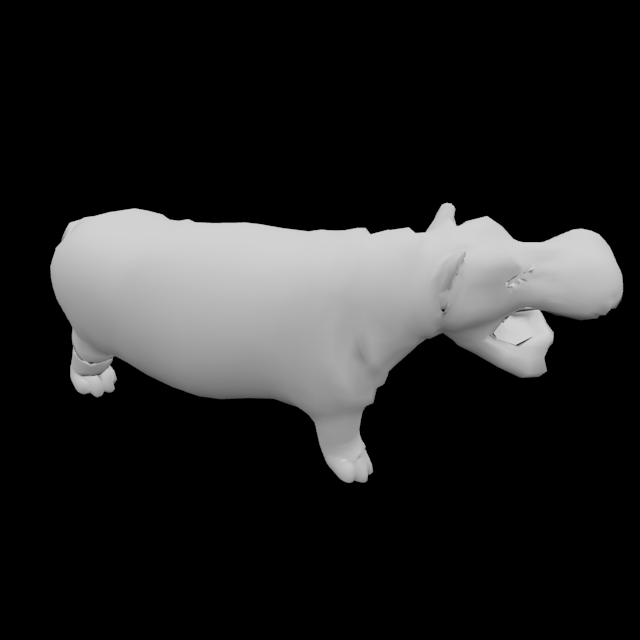}
        \hspace{0.01\linewidth}
        \includegraphics[width=0.3\linewidth]{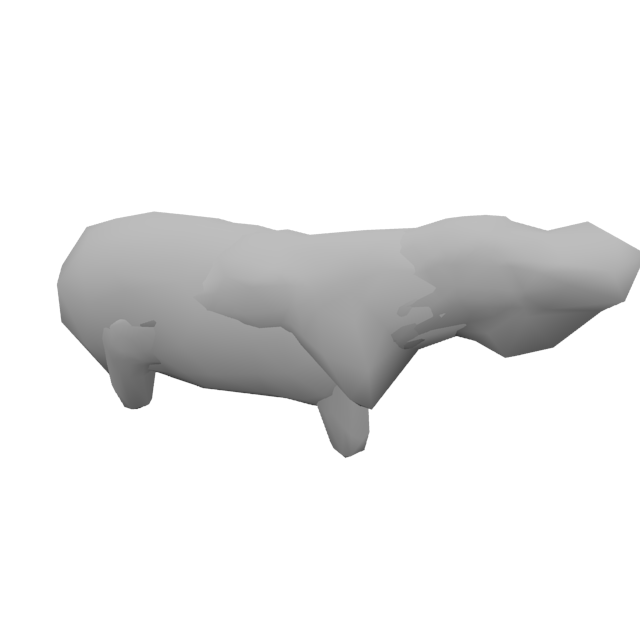}
        \hspace{0.01\linewidth}
        \includegraphics[width=0.3\linewidth]{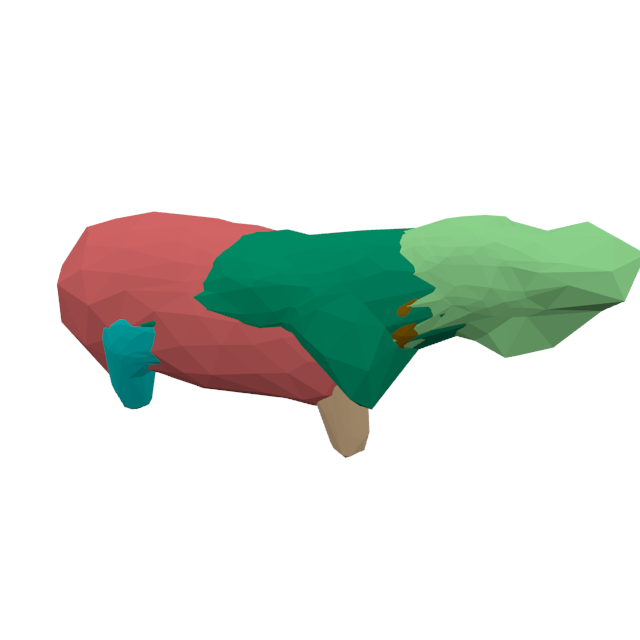}
        \caption{Hippo}
        \label{fig:hippo}
    \end{subfigure}
    \caption{3D Animal Reconstruction. We visualize $4$ examples from the test split and the 3D outputs of Cerberus. \textbf{Left:} Input images. \textbf{Middle:} 3D outputs. \textbf{Right:} Parts rendered in different colors.}
    \label{fig:animal}
\end{figure}

We demonstrate the predicted 3D animals and parts from Cerberus in Figure~\ref{fig:animal}. Based on the visualized test outputs, we see that Cerberus predicts high quality meshes for different animals in various poses from diverse viewpoints. Additionally, part segmentation is reasonable and consistent across different animals, but parts vary appropriately in shape. For example, the light green part always corresponds to the head, but this part has a pointed snout for the deer in Figure~\ref{fig:deer} and a round snout for the hippo in Figure~\ref{fig:hippo}.

\section{Conclusion}
We have proposed a new architecture and training paradigm for single-image 3D reconstruction with only 2D supervision. Our approach not only reconstructs 3D models more accurately than approaches that use a single monolithic mesh, but also infers semantic parts without part-level supervision.

Although we focus on the problem of 3D reconstruction, in the spirit of inverse graphics, our approach can potentially be adapted to tasks such as classification or pose estimation. Current state-of-the-art approaches to these tasks rely on extensive amounts of labeled training data. A 3D representation that explicitly disentangles shape, pose, and viewpoint has the potential to significantly improve sample efficiency, because the basic invariance properties of objects are reflected directly in the representation and need not be learned from labels.

{\small
\bibliographystyle{ieee}
\bibliography{references}
}

\end{document}